\documentclass[conference]{IEEEtran}
\IEEEoverridecommandlockouts
\usepackage{cite}
\usepackage{amsmath,amssymb,amsfonts, amsthm, subcaption}
\usepackage{graphicx}
\usepackage{tabularx}
\usepackage{textcomp}
\usepackage{xcolor}

\theoremstyle{definition}
\usepackage{pifont}
\newcommand{\cmark}{\ding{51}}%
\newcommand{\xmark}{\ding{55}}%
\def\BibTeX{{\rm B\kern-.05em{\sc i\kern-.025em b}\kern-.08em
    T\kern-.1667em\lower.7ex\hbox{E}\kern-.125emX}}
\begin{document}
\bibliographystyle{ieeeTran}

\title{Physics Informed Machine Learning (PIML) methods for estimating the remaining useful lifetime (RUL) of aircraft engines\\
\thanks{Work performed in Southwest Research Institute (SwRI)}
}

\author{\IEEEauthorblockN{Sriram Nagaraj}
\IEEEauthorblockA{\textit{Southwest Research Institute (SwRI)} \\
San Antonio, Texas, USA \\
sriram.nagaraj@swri.org}
\and
\IEEEauthorblockN{Truman Hickok}
\IEEEauthorblockA{\textit{Southwest Research Institute (SwRI)} \\
San Antonio, Texas, USA \\
truman.hickok@swri.org}

}

\maketitle

\begin{abstract}

This paper is aimed at using the newly developing field of physics informed machine learning (PIML) to develop models for predicting the remaining useful lifetime (RUL) aircraft engines. We consider the well-known benchmark NASA Commercial Modular Aero-Propulsion System Simulation (C-MAPSS) data as the main data for this paper, which consists of sensor outputs in a variety of different operating modes. C-MAPSS is a well-studied dataset with much existing work in the literature that address RUL prediction with classical and deep learning methods. In the absence of published empirical physical laws governing the C-MAPSS data, our approach first uses stochastic methods to estimate the governing physics models from the noisy time series data. In our approach, we model the various sensor readings as being governed by stochastic differential equations, and we estimate the corresponding transition density mean and variance functions of the underlying processes. We then augment LSTM (long-short term memory) models with the learned mean and variance functions during training and inferencing.  Our PIML based approach is different from previous methods, and we use the data to first learn the physics. Our results indicate that PIML discovery and solutions methods are well suited for this problem and outperform previous data-only deep learning methods for this data set and task. Moreover, the framework developed herein is flexible, and can be adapted to other situations (other sensor modalities or combined multi-physics environments), including cases where the underlying physics is only partially observed or known.

\end{abstract}


\section{Introduction}

Aircraft maintenance is a a critical component in ensuring the safety, dependability, and efficiency of aviation systems and their operational modes (\cite{huan2020fault, singh2020review, tian2020health, tong2018review}). Classical maintenance methods, such as scheduled maintenance, aimed at corrective repair/upkeep may still be relevant but are often plagued by limitations such as operational (including logistic) disruptions, periods of unavailability (downtime) and increased costs, for even routine issues. On the other hand, predictive maintenance (PM), which refers to maintenance approaches relying on data-driven or data-augmented techniques, is a proactive set of approaches that leverage past data (and modes) to foresee system (or subsystem) failures prior to occurrence (\cite{wang2019data, yang2019condition}). Quantities of interest that may be part of a PM diagnosis are probability of engine failure/stalls, estimating remaining useful lifetime (RUL) of engines, probability of catastrophic multi-system failure etc. A characteristic of PM systems are real-time (or near real-time) information about the system and dynamic updating of data to provide estimates when the mission or flight is in progress. Moreover, these systems can be incorporated into the flight system itself and not only in separate command centers.
\subsection{Salient Features:}Other important points to note about PM are:
\begin{itemize}
\item Since PM systems are (near) real-time, they allow for early detection and identification of aircraft degradation or faulty functioning of aircraft subsystems. In the case of commercial aircraft, this allows to potentially save passenger lives by reducing the risk of dangerous malfunction being unnoticed. PM allows for corrective, risk-mitigation maneuvers on the part of pilots (\cite{wang2019data}).
\item When early onset of faulty system performance is picked by PM diagnosis, this allows for action to be taken that would help avoid further catastrophic outcomes such as complete engine failure. This would help not only saving human life, but also reduce total maintenance costs of the mission/flight (\cite{tian2020health, tong2018review}).
\item PM allows for addressing maintenance needs during both scheduled and unscheduled periods. This in turn minimizes any unplanned downtime of aircraft systems and helps to optimize aircraft utilization.
\item Continuing the earlier remarks, early detection of system malfunction allows for the use of maintenance resources for mission-critical tasks. Thus, PM ensures the focusing of in-depth maintenance efforts on aircraft/avionics systems or components that are known to require attention, thereby optimizing the allotment of maintenance resources. Finally, this helps to reduce costs overall in maintenance.
\item As another means of reducing overall costs, PM reduces the chances of unexpected maintenance issues, and thereby reduces operational delays (such as flight delays, cancellation or diversions) which are known to impact commercial airlines significantly. For military aircraft, operational delays may impact the efficacy of the mission, and may even lead to a critical mission being compromised (\cite{wang2019data}).

\item PM and ML/AI models: PM relies on innovative use of sensor data from various aircraft systems, such as engines, avionics, and airframes. These sensor readings are fused via ML/AI techniques and processed to predict anomalous behavior, potential known or unknown failures; further, in non-critical scenarios, the ML/AI methods are used to monitor equipment health. Simply put, the ML/AI techniques are trained on a large swathe of past sensor data in order to detect/identify trends and risk factors that are indicative of equipment failures or system degradation. Once detected or identified, information about these issues can be communicated to a subject matter expert (SME) to determine the severity of the issues and what corrective actions need to be taken (\cite{singh2020review}).

\item Regulatory agencies such as the Federal Aviation Administration (FAA) and the European Union Aviation Safety Agency (EASA), have developed standards and regulations that need to be met by commercial aircraft in order for them to be flight worthy. PM techniques can be used to augment existing risk management methods to further ensure compliance with these regulatory standards.

\end{itemize}
There are several challenges to full-scale incorporation of PM methods. First, data quality/integrity and interoperability challenges often prevent the implementation of PM methods. Scarce data or uneven data between different sensors limit the applicability of PM techniques. Integrating real-time systems on platform is a separate challenge, as the existing hardware and infrastructure on aircraft needs to be sufficient for running PM diagnostics. Finally, reliability of PM methods requires extensive testing, and both commercial and military operators need to allocate sufficient resources to adequate testing. However, these challenges can be surmounted with careful analysis and a well-defined systems engineering approach during design and development. The future directions of a PM based fleet are to include internet of things (IoT) devices into PM data sources and the design of digital twins to further improve the efficacy of PM.

PM is therefore an emerging, and continually evolving paradigm in aircraft maintenance practices (\cite{huan2020fault}). PM techniques can be incorporated with other evolutions in the aviation industry with newer hardware/sensing capabilities. PM can then co-evolve to ensure the safety and reliability of ever increasing complex aircraft systems. Moreover, the financial savings afforded by PM can be used to further increase aircraft capabilities.
\subsection{Physics Informed Machine Learning}
Currently, a large amount of computational modeling is based on deterministic models and/or multi-fidelity approximate solutions that do not directly account for inherent variability (stochasticity) in the material properties. When predicting real-world expected performances, these models are laboriously recomputed many times and deliver solutions that do not generalize efficiently to situations where the data needed for simulations is partially observed. Moreover, real experimental test data is a scarce commodity, so there is a need to utilize such data in the most efficient way possible (\cite{kutz2nd}).
\subsubsection{PIML in a Nutshell}
Physics Informed Machine Learning (PIML) models offer a promising way forward for balancing computational fidelity and data-driven generalization. PIML is a data driven ML modeling framework that is augmented by the underlying scientific domain (mechanics, electromagnetics, acoustics, thermal or multiphysics). Thus, scientific knowledge captured by some relevant set of physical principle/equation (conservation laws, Maxwell equations, mechanical properties, heat diffusion etc.) is directly integrated with the ML modeling to predict quantities of interest (failure of materials, energy transmission efficiency, stress, drag etc.) which are often needed for mission-critical decision making (\cite{kutz2nd,raissi2019physics}).
PIML models achieve better generalization than traditional “data only” ML models since they can be viewed as incorporating real-world prior knowledge (i.e. a Bayesian reformulation). This also allows PIML models to operate in low/partially observed data regimes where the physics-informed component can act as data surrogate as needed. Due to inherent regularization, PIML models are less prone to spurious solutions while dealing with noisy data as well. Finally, PIML models yield more succinct models characterized by fewer ML parameters while achieving (often exceeding) the accuracy/efficiency of complex “data only” deep learning models. This also renders then more explainable than “black box” ML models and is significant when downstream decision making requires explainable predictions (\cite{xu2020physics,zhang2020physics}).
\subsubsection{Methodologies in Physics-Informed Machine Learning} There are several ways in which PIML performed.
\begin{itemize}
\item Differential Equation Constraints: PIML frameworks utilize differential equations to incorporate dynamic system behavior into machine learning models, facilitating time-series forecasting, and dynamical system identification (\cite{zhang2020physics}).
\item Generative Data: Systems governed by stochastic differential equation (SDE) models can be used to perform physics-informed data generation. This generated data would capture well the underlying physics, and being a data augmentation method, can directly be incorporated into the ML modeling to ensure model robustness (\cite{yang2020physics}). 
\item Hybrid Physics-Data Models: Hybrid approaches are also possible which integrate physics-based simulations with data-driven (ML) techniques, balancing model accuracy and computational efficiency (\cite{xu2020physics}).
\end{itemize}
\subsubsection{Modes of PIML Operation}
Roughly, PIML methods operate in two modes. The first is the ``discovery mode" wherein the physical parameters (coefficients of differential equations, structural parameters etc.) are learning from the underlying data. The second is the ``solution mode" which uses a fully specified physical model along with a data-driven component to drive predictions. The exact means by which the two modes operate are largely problem/domain dependent, and specific implementations of either mode of operation may change widely across different areas (\cite{yang2020physics,raissi2019physics}).

\subsubsection{Physics Informed Neural Networks (PINNs)}  PINNs are an exemplary class of PIML models. PINNs are deep neural networks which are trained on a loss functions composed of two components (\cite{raissi2019physics}). The first is the usual data driven (often mean square error MSE), while the other component is a physics-informed one. The physics informed loss includes all physical constraints of the problem include modeling choices, differential/integral equations, boundary conditions, uncertain coefficients etc. Both components of the overall loss work in tandem to deliver a robust model. While PINNs are one example of a class of PIML models, it is possible to have other PIML approaches as well. In this regard, \emph{PIML is a modeling framework, rather than a specific set of model types}.
\subsubsection{Applications of PIML}
\begin{itemize}
\item Aerospace Engineering: PIML methods are very useful for turbulence modeling, wing design and shape optimization, where traditional computational methods such as finite element/finite difference methods can be augmented by data-driven models (\cite{ghavamian2020physics,xu2020physics}).
\item Mechanical Engineering: Applications in mechanical engineering includes modeling of fatigue of materials, CFD, elasticity etc. (\cite{ling2016reynolds,wang2020physics,xu2020physics})
\item Material Science: PIML accelerates materials discovery by integrating physics-based simulations with high-throughput experimentation, guiding the design of novel materials with desired properties (\cite{xu2020physics,piml_material_science}).
\item Biomedical Engineering: PIML aids in medical image analysis, patient monitoring, and drug discovery, leveraging physics-based models and clinical data to improve diagnosis and treatment outcomes (\cite{zhu2021physics,hamilton2021learning,onieva2020physics}).
\end{itemize}
Scalability of PIML methods to high-dimensional data remains a challenge. Lower order modeling, model order reduction methods, and surrogate models have proven to be promising ways of mitigating this challenge (\cite{zhu2020physics}). Another area requiring further investigation is uncertainty quantification (UQ) of PIML predictions. UQ is needed for meaningful decision-making and risk assessment, and robust methods for uncertainty quantification and propagation are essential. Finally, software packages for easy implementation of differential equation solvers using deep learning methods are available as well (\cite{lu2019deepxde}).

\subsection{Prior Work: Data-Driven Deep Learning for PM using the C-MAPSS Dataset}
In recent years, machine learning techniques, particularly long short-term memory (LSTM) and convolutional neural networks (CNNs), have shown promise in PM applications. Towards this end, the Commercial Modular Aero-Propulsion System Simulation (C-MAPSS) dataset (a widely used benchmark dataset for prognostics and health management research) is a standard dataset used for studying the efficiency of ML models for PM (\cite{saxena2008damage,si2011prognostics}). C-MAPSS consists of multiple simulated engine run-to-failure trajectories, each containing sensor measurements and corresponding remaining useful life (RUL) labels. The data consists of multiple noisy time series that correspond to sensor readings from different parts of an aircraft engine. Thus far, the LSTM Networks and CNNs have been well-studied architectures for PM usinng the C-MAPSS dataset (\cite{lstmrul, jiang2020hybrid,yang2018deep,zhou2018remaining}). LSTMs are a type of recurrent neural network (RNN) designed to capture long-term dependencies in sequential data. LSTMs consist of memory cells, input and forget gates, and output gates, allowing them to retain information over long sequences and adapt to varying temporal patterns. CNNs on the other hand, are deep learning architectures designed for feature extraction from structured data, such as images or time-series signals. CNNs consist of convolutional layers, pooling layers, and fully connected layers, enabling hierarchical feature learning and pattern recognition. CNNs are applied to sensor data for fault detection, anomaly detection, and condition monitoring, leveraging their ability to automatically learn relevant features from raw data. LSTMs have been successfully applied to time-series data for equipment health monitoring, remaining useful life estimation, and fault diagnosis. CNNs have been applied to sensor data for fault detection, anomaly detection, and condition monitoring, leveraging their ability to automatically learn relevant features from raw data.
However, these methods are purely data-driven and do not incorporate the physics underlying the data.
\begin{figure}
\centering
\begin{subfigure}[b]{0.55\textwidth}
   \includegraphics[width=1\linewidth]{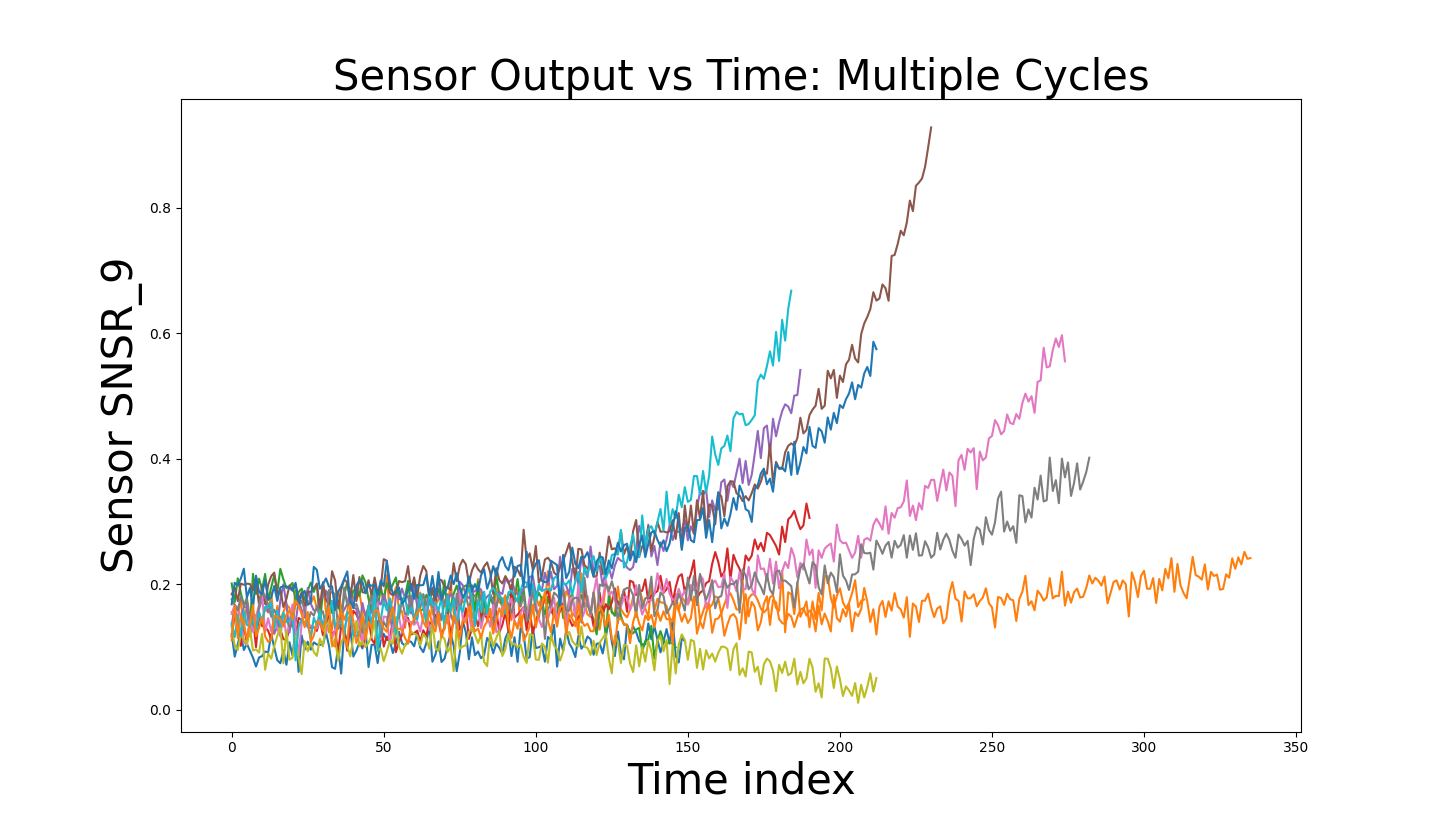}
   \caption{}
   \label{fig:fig1} 
\end{subfigure}

\begin{subfigure}[b]{0.55\textwidth}
   \includegraphics[width=1\linewidth]{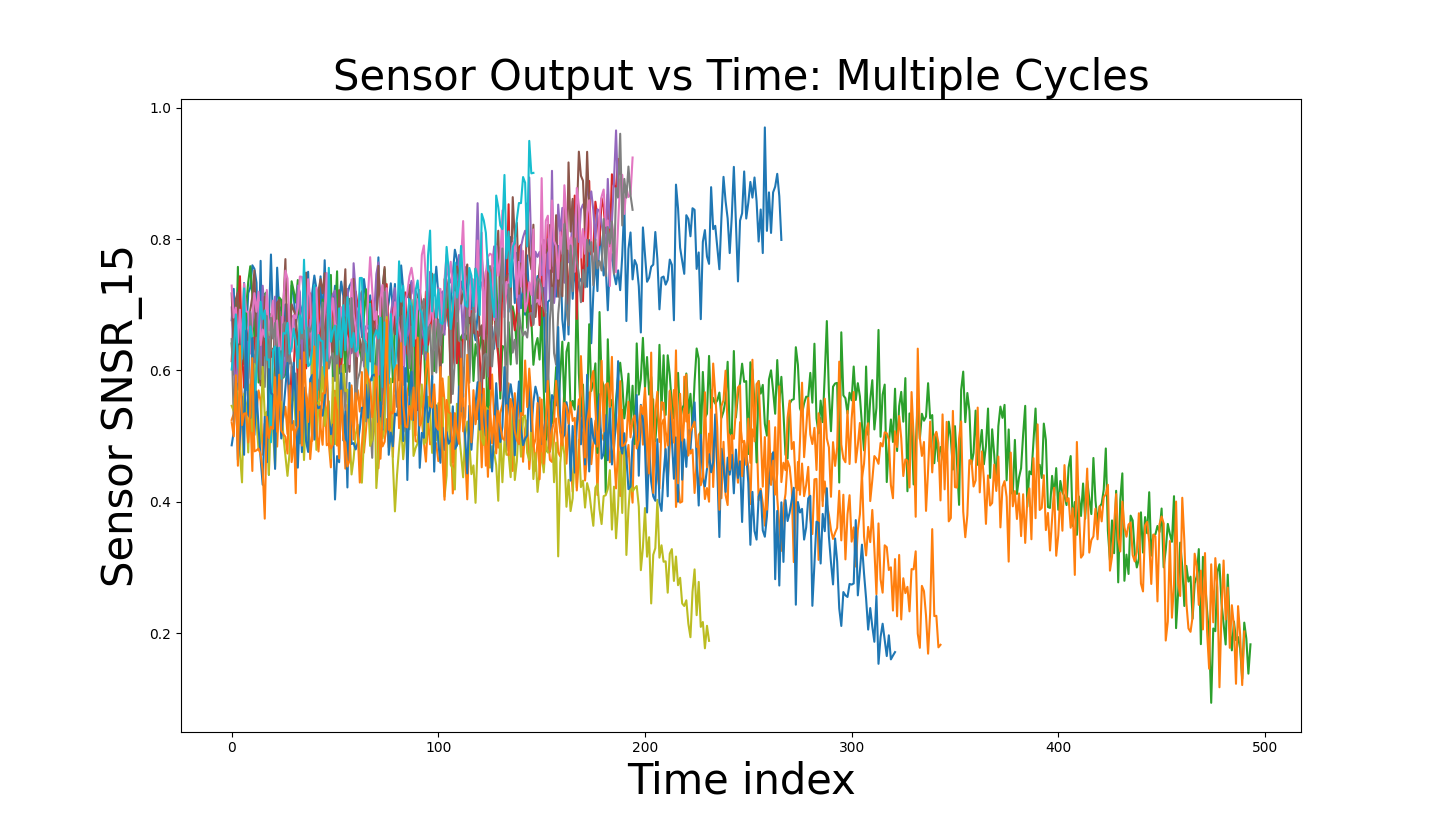}
   \caption{}
   \label{fig:fig2}
\end{subfigure}

\caption[Two numerical solutions]{(a,b) Sample sensor outputs for two different modes of operations (FD001 and FD003 respectively). (a) is well modeled as geometric brownian motion, while (b) shows multi-modal behavior of sensor output.}
\end{figure}

\subsection{Our Contributions:}
\begin{figure}
\centering
\begin{subfigure}[b]{0.55\textwidth}
   \includegraphics[width=1\linewidth]{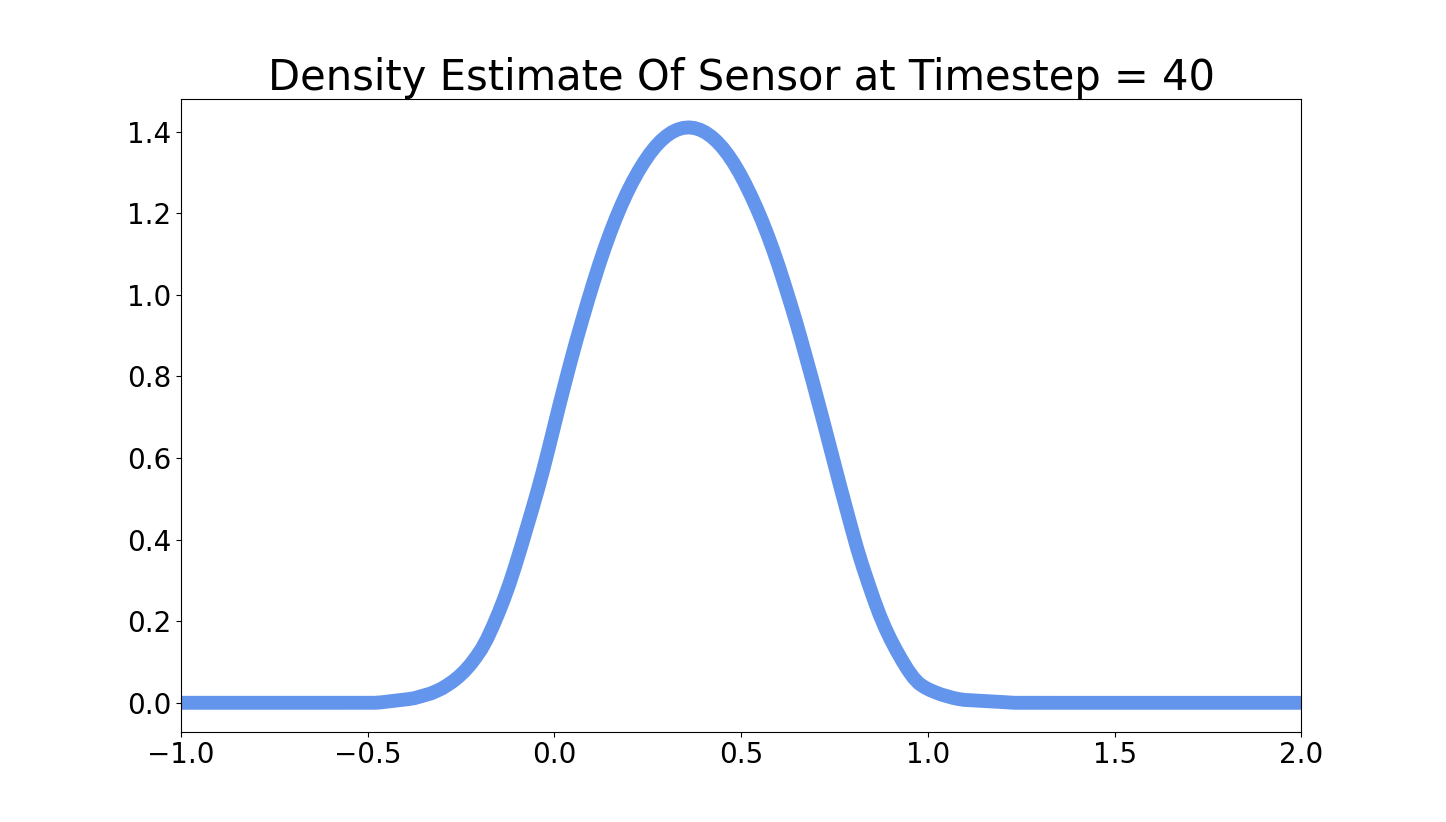}
   \caption{}
   \label{fig:fig3} 
\end{subfigure}

\begin{subfigure}[b]{0.55\textwidth}
   \includegraphics[width=1\linewidth]{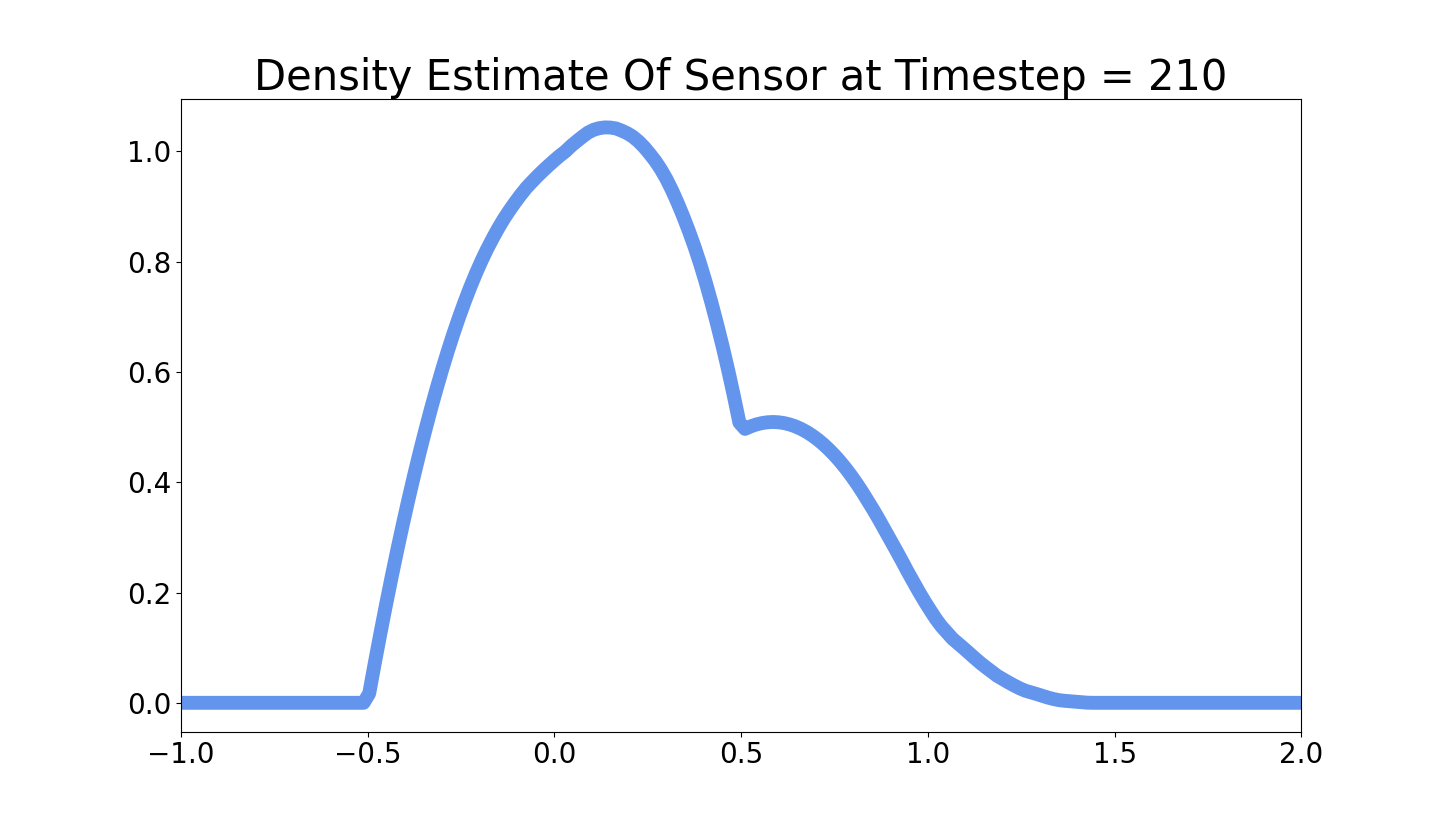}
   \caption{}
   \label{fig:fig4}
\end{subfigure}

\caption[]{(Evolution of the histogram over time for the same sensor. Notice the shift from unimodal to bimodal behavior.}
\end{figure}

We consider the C-MAPSS dataset as the main data for this paper. As we have seen, this is a well-studied dataset, and much work exists in the literature that address RUL prediction on C-MAPSS data with classical and data-driven (deep) ML methods. Our PIML based approach is different from previous methods, and we use the data to first learn the physics. In the absence of published empirical physical laws governing the C-MAPSS data, our approach first uses stochastic PIML methods to estimate the governing stochastic physics models from the noisy time series data (this is referred to as the ``discovery” mode of PIML). In this phase, we model the sensor data as being governed by a system of first order stochastic differential equation (SDE) and estimate the mean and variance functions. When the underlying statistical distribution at a given timestep is known to be unimodal, we can use the sample mean and variance as estimates of the mean and variance functions at that timestep, or equivalently, the estimate of the $K=1-$mean cluster median and variance. If the distribution is multimodal (with $K-$modes) we use $K$-means clustering estimation to define the estimated mean and variance functions at the timestep as the $K-$medians of the collection of sample paths at the timestep in question. We are thereby able to obtain a surrogate for the underlying physics. $K$-means estimation is a well-established nonparametric method, and using it, we are able to model multi-modal sensor output wherein the statistical distribution of the same sensor output is a multi-modal distribution (see Figure \ref{fig:fig2} and Figure \ref{fig:fig4}). We then use existing ML models (LSTM) with the learned physics to generate predictions of RUL (the ``solution” mode of PIML). Unlike traditional PIML methods, in the solution mode, we directly incorporate the mean and variance functions as data augmentation rather than indirectly through a loss function. The proposed set of methods delivers reliable, robust solutions to the challenging problems of physics discovery and physics informed prediction, relying on the superior generalization capabilities of modern deep machine learning. Our results indicate that PIML discovery and solutions methods are well suited for this problem and outperform previous data-only deep learning methods for this data set and task. Moreover, the framework developed herein is flexible, and can be adapted to other situations (other sensor modalities or combined multi-physics environments), including cases where the underlying physics is only partially observed or known. 

\section{Stochastic Physics Model}
We now describe our physics model for the different datasets. Recall the absence of existing physics based models for the data. We therefore appeal to the approach of \emph{learning} the physics from the raw data. To this end, we model the output $S(t)$ of each (viable, information bearing) sensor via a stochastic differential equation (SDE). Note that, inspecting the sample paths of the sensors from FD001-FD004 (see section \ref{sec:exp} for more details on the dataset) shows that this is a reasonable modeling choice. Indeed, several sensors are well-modeled as geometric brownian motion. We assume the following form for each sensor output $S:$

\begin{equation}dS(t) = a(t)dt + b(t)dW(t), \end{equation}
where $W(t)$ is a one-dimensional martingale stochastic process (not necessarily Gaussian, but with independent increments) and $a(t),b(t)$ are the \emph{drift and diffusion} coefficients respectively. The stochastic integrals are assumed to be in the sense of Ito. Since $S(t)$ is a stochastic process, we will, as needed, explicitly denote the dependence on a random outcome in the event space $\omega$ as $S(t,\omega)$. In reality, the drift and diffusion functions are themselves dependent on both $t$ as well as $S(t),$ i.e., $a(t) = \tilde{a}(t,S(t))$ and $b(t) = \tilde{b}(t,S(t))$, although for our purposes, we shall denote them as functions of $t$ alone, with the dependence on $S(t)$ being implicit. Thus, each data set consists of multiple cycles of several sensor values, $S_{i}(t)$ with $i=1,\ldots,N.$ One possible approach would be to estimate the corresponding $a_i(t), b_i(t),$ from the data directly to fully specified the SDE model. However, we take a slightly different approach. We instead estimate the mean and variance process defined by:

\begin{align}\mu(t)&= \mathbb{E}[S(t)],\\\rho(t) &= \mathbb{E}[|S(t)-\mu(t)|^2].\end{align}
These are the instantaneous mean and variance functions of $S(t)$. Note that if $p(t,x)$ denotes the density of the process $S(t,\omega),$ i.e., $S(t,\cdot)\sim p(t,x)$ we have that $\mu(t) = \int xp(t,x) dx$ and $\rho(t) = \int (x-\mu(t))^2p(t,x)dx.$ We take this approach for the following reasons. First, an inspection of sensor data as a function of time (over multiple cycles) indicates that several sensors have non-Gaussian distribution at different time instances (see for e.g. Figure \ref{fig:fig2}). Typical use cases of the standard diffusive Ito SDE assume Gaussian, independent increment processes, which would not be valid in the current case. Second, estimating the coefficients $a(t),b(t)$ directly from data would need further assumptions on the underlying process $S(t),$ which, again, may not be valid. Instead, we stay with the most general assumptions possible, and estimate the mean and variance functions instead. Here, the only assumption being made is that the process has finite first and second moments, which is evidently true.
\subsection{Estimating $\mu(t), \rho(t)$}
We now discuss the estimation of $\mu(t), \rho(t)$. We first assume that the distribution of $S(t,\omega)$ is unimodal for all $t$. This is true for several sensors in the dataset. For a given sample path $S_{i}(t,\omega_k),$ we define the support of the path as \begin{equation}\rho_{S_{i}}(k):=\{t>0:|S_{i}(t,\omega_k)|>0\}.\end{equation}
We assume that $\rho_{S_{i}}(k) = [0,T_{S_{i}}(k)]$ for some $T_{S_{i}}(k)>0.$ We set $T_{max} = \text{sup}_{\omega_k,S_{i}}(T_{S_{i}}(k)) $ to be the maximum length of the vector of any sensor's readings over all cycles. If a given sample path $S_{i}(t,\omega_k)$ ends before $T_{max}$, we extend the sample path by zero, i.e., we set $S_{i}(t,\omega_k) = 0, T_{S_{i}}(k)<t<T_{max}.$

Now, given a fixed partition $\Delta = \{0<t_1,\ldots,t_n = T_{max}\},$ for each $t_k$ we define a random variable as $\tilde{S_i}(k,\omega) = S_i(t_k,\omega)$. That is, $\tilde{S_i}(k,\cdot)$ is obtained by collecting the sample paths $S_i(t,\cdot)$ and evaluating them at $t=t_k,$ thereby resulting in the random variable $\tilde{S_i}(k,\cdot):\Omega \rightarrow \mathbb{R}.$ We then define $\hat{\mu}_i(k) = \mathbb{E}(\tilde{S_i}(k,\cdot)),$ and $\hat{\rho}_i(k) = \mathbb{E}(\tilde{S_i}(k,\cdot)^2)-\hat{\mu}_i(k)^2$ to be the mean and variance of the random variable $\tilde{S_i}(k,\cdot).$ Finally, we can define the estimates of the drift and diffusion functions via an interpolation operator $\Pi: \mathbb{R}^n \rightarrow C([0,T_{max}])$ as $\hat{\mu}_i(t) = \Pi(\{\hat{\mu}_i(k),k=1,\ldots,n\})$ and likewise $\hat{\rho}_i(t) = \Pi(\{\hat{\rho}_i(k),k=1,\ldots,n\}).$

In practise, we shall be in need of only the values $\hat{\mu}_i(k),k=1,\ldots,n$ (resp. $\hat{\rho}_i(k)$), and hence will not specify the exact form of the interpolation operator $\Pi(\cdot).$ Indeed, we preprocess the sensor data and compute the drift and diffusion functions using an appropriately defined histogram (or density estimate) of the random variable $\tilde{S_i}(k,\omega)$ at each time instant $t_k\in \Delta.$ Computing $\hat{\mu}_i(k)$ is done using $K-$ means clustering. Indeed, since we know the distribution is unimodal, we compute the single centroid of the distribution of $\tilde{S_i}(k,\cdot)$ and define it to be the mean $\hat{\mu}_i(k)$. This way, we avoid explicit computation of the expectation.

\subsubsection{Estimating $\mu(t), \rho(t)$: Multimodal Case}
Thus far, we have tacitly assumed that the histogram (density estimate) of $\tilde{S_i}(k,\omega)$ is unimodal when computing the drift and diffusion functions. In some cases, this is not true, and we need to modify our approach to include multi-modal densities. In such cases, we resort to modeling the multi-modal density of $\tilde{S_i}(k,\omega)$ using a $K-$means estimate of the mean. Typically, there are not more than two components in the mixure. Accordingly, the same approach as earlier can be carried forward (using $K=2$), with the caveat that we now have two mean function estimates $\hat{\mu}^{1}(t),\hat{\mu}^{2}(t)$ (resp. $\hat{\rho}^{1}(t),\hat{\rho}^{2}(t)$). The particular sensors that exhibit this behavior can be ascertained beforehand, and this multi-modal analysis can be done only in such cases.  
\subsection{Generative Modeling: Synthetic Data Generation}
\begin{figure}
\centering
\begin{subfigure}[b]{0.55\textwidth}
   \includegraphics[width=1\linewidth]{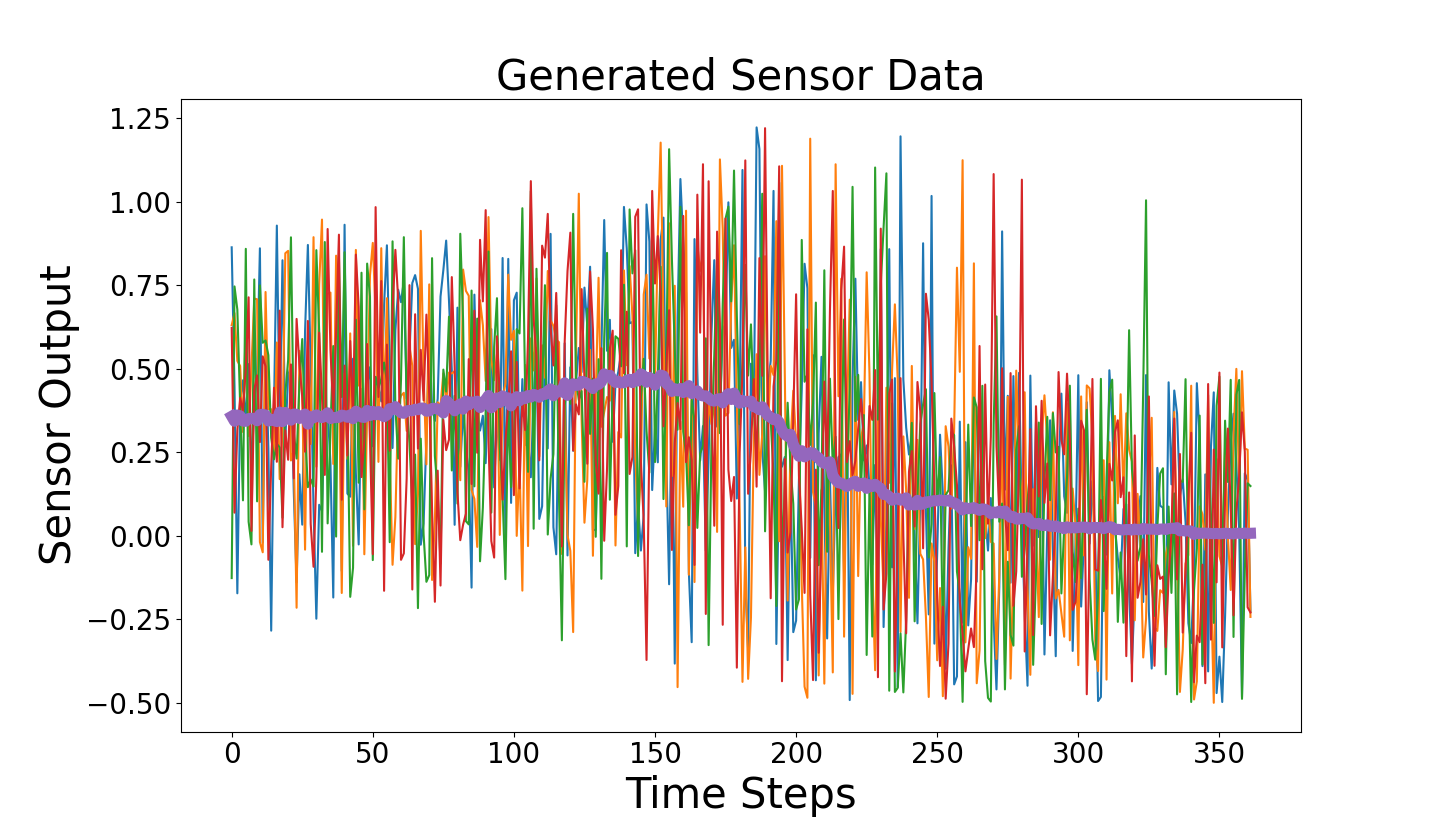}
   \caption{}
   \label{fig:fig5} 
\end{subfigure}

\begin{subfigure}[b]{0.55\textwidth}
   \includegraphics[width=1\linewidth]{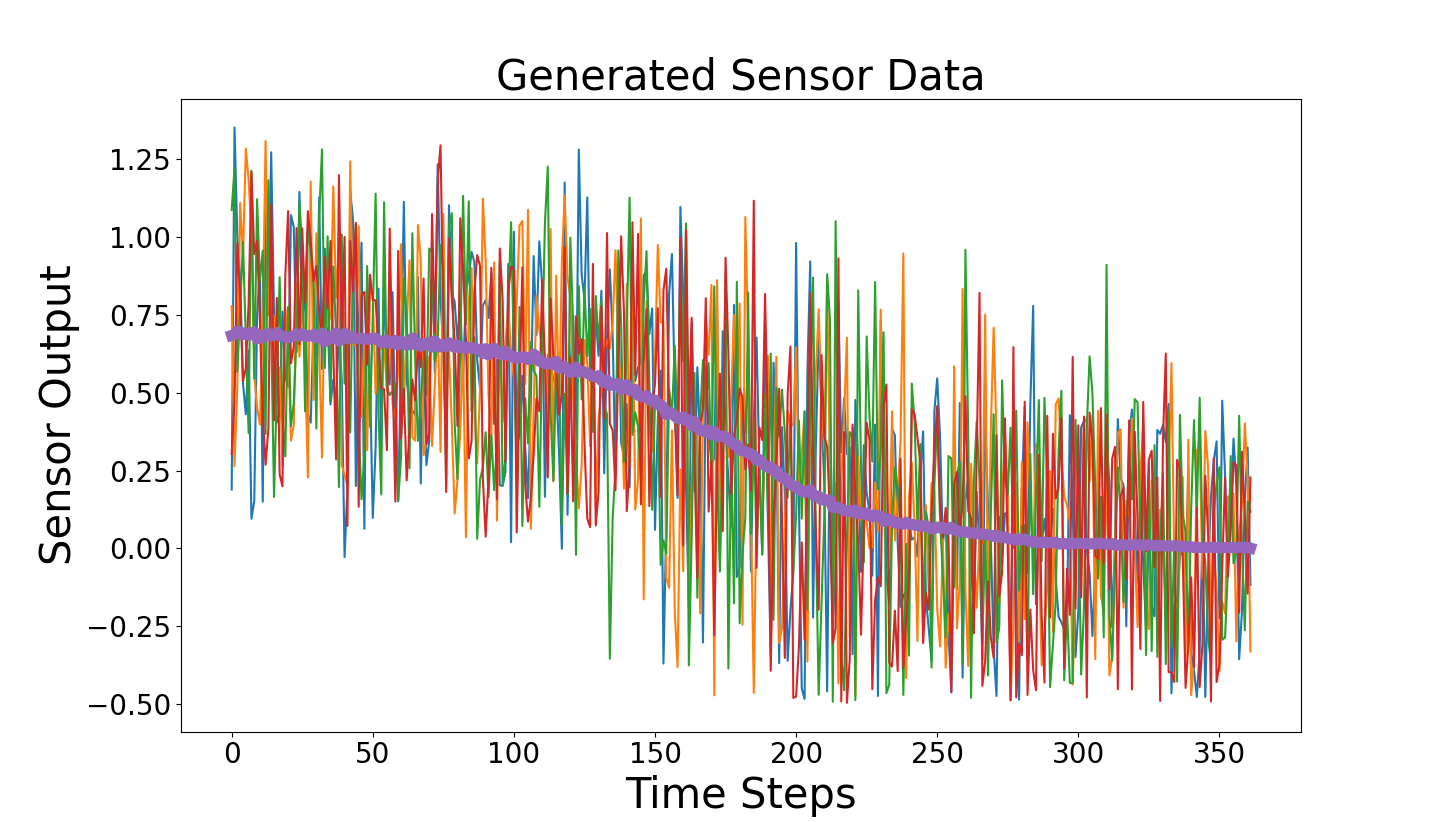}
   \caption{}
   \label{fig:fig6}
\end{subfigure}

\caption[]{Sample generated (synthetic) data from the estimated distributions. In bold is the mean of the actual (real) data.}
\end{figure}

Note that the above approach also yields a \emph{generative} model for information-bearning sensors as well. Indeed, in the standard diffusive SDE case, assuming that $W(t)$ is a Brownian motion, we know from standard SDE theory that the density $p(x,t)$ of $S(t)$ is a solution of the Fokker-Planck equation: \begin{equation}\frac{\partial }{\partial t}p(x,t) = -\frac{\partial }{\partial x}\mu(x,t)p(x,t) + \frac{\partial^2 }{\partial x^2}\rho(x,t)p(x,t).\end{equation}
In our current case, we do not have such a clear mathematical form for the evolution of the density. However, we can sample from the $K$-means density estimate that we used to compute the mean process at each time instant $t_k\in\Delta$ to obtain generated sample paths. This is a unique feature of our approach, and yields a physics-informed synthetic data generation mechanism that can be used to augment the raw (real) sensor data. Figures \ref{fig:fig5} and \ref{fig:fig6} show how we can generate samples from the time-varying densities which may possibly be multi-modal. The samples that are generated evidently have similar mean/variance properties as the ``real" data. Notice that, since $dW(t)$ is assumed to be a zero-mean process, we trivially have \begin{align}\frac{d\mu(t)}{dt} &= \bar{a}(t), \\\frac{d\rho(t)}{dt} &= \frac{dR(t)}{dt}-2\mu(t)\bar{a}(t), \end{align}
where $\bar{a}(t) = \mathbb{E}[a(t)], \, R(t) = \mathbb{E}[S(t)^2].$

\section{Experiments}\label{sec:exp}
 
\subsection {Dataset}
We focus our experiments on the aforementioned C-MAPSS dataset produced by NASA, which is organized into four subsets corresponding to distinct aircraft operating conditions. Each subset contains synthetic sensor readings meant to simulate run-to-failure trajectories of turbofan engines. Each trajectory is usually a few hundred timesteps. We drop 8 of the 22 available sensors (sensors 1, 5, 6, 10, 16, 18, 19, and 22) and split each trajectory into 20-timestep windows, which are shuffled during training.
 
\subsection {Deep Learning Model}
 
Consistent with previous work \cite{lstmrul}, we use a small LSTM neural network to predict the RUL at the end of each 20-timestep window. Specifically, we use a single-layer LSTM with 12 hidden units and apply a two-layer MLP to the LSTM's hidden state after all 20 timesteps have been processed.
 
We train on each of the four operating conditions in isolation, which aligns with real-world applications. We train the model to minimize mean squared error (MSE) of the RUL at the end of each 20-timestep window:
 
\begin{equation}
    J = \frac{1}{B}\sum_{B} \left\lVert RUL^{pred}_{t} - RUL^{gt}_{t} \right\rVert^2
\label{eq:rul_loss}
\end{equation}
 
where $B$ is the batch size (64 in our case), $RUL^{pred}_{t}$ is the RUL predicted by the model at timestep $t$ (end of the 20-timestep window), and $RUL^{gt}_{t}$ is ground truth. During testing, we predict the RUL at the end of each trajectory in the test set and report MSE across the entire test set (same as Equation~\ref{eq:rul_loss}) along with the average L1 error. We report the average of these metrics across 3 random seeds.
 
We use an Adam optimizer and a learning rate of 0.001. We train the model for 100 epochs on the FD001 and FD003 subsets and for 1000 epochs on the FD002 and FD004 subsets. We also used early stopping, testing the best-performing model from any epoch.
 
\subsection {Results}
 
Table~\ref{tab:combined_table} presents the MSE and L1 error of the model under each operating condition. The leftmost columns indicate the operating condition and whether $\mu$ and/or $\rho$ were included as input during each training run.
 
Under all four operating conditions, including $\mu$ \textit{and} $\rho$ significantly outperforms the models where neither $\mu$ or $\rho$ are included. $\rho$, however, appears to contain less information than $\mu$, as shown by our $\rho$-only ablations; these runs nonetheless outperformed the baseline under all operating conditions. FD002 and FD004 are clearly of much greater complexity than FD001 and FD003, which is consistent with previous work \cite{lstmrul}.  
 
\begin{table}[h]
    \centering
    \begin{tabular}{ccccc}
    \textbf{Condition} & \textbf{Mu} & \textbf{Rho} & \textbf{Test MSE} & \textbf{Test L1} \\
    FD001 & \xmark & \xmark & 4.73 & 1.21 \\
    FD001 & \cmark & \xmark & 1.24 & 0.55 \\
    FD001 & \xmark & \cmark & 2.13 & 0.91 \\
    FD001 & \cmark & \cmark & 1.02 & 0.59 \\
    \hline
    FD002 & \xmark & \xmark & 430.44 & 15.41 \\
    FD002 & \cmark & \xmark & 309.23 & 13.15 \\
    FD002 & \xmark & \cmark & 282.44 & 12.62 \\
    FD002 & \cmark & \cmark & 277.66 & 12.52 \\
    \hline
    FD003 & \xmark & \xmark & 3.47 & 1.40 \\
    FD003 & \cmark & \xmark & 1.69 & 0.42 \\
    FD003 & \xmark & \cmark & 3.32 & 0.57 \\
    FD003 & \cmark & \cmark & 0.15 & 0.29 \\
    \hline
    FD004 & \xmark & \xmark & 944.21 & 22.50 \\
    FD004 & \cmark & \xmark & 670.47 & 19.05 \\
    FD004 & \xmark & \cmark & 720.76 & 19.57 \\
    FD004 & \cmark & \cmark & 659.21 & 18.79 \\
    \end{tabular}
    \caption{Test set mean squared error and L1 error under different operating conditions (FD001, FD002, FD003, FD004), according to whether mu and rho were added as inputs. Results are averaged across 3 seeds.}
    \label{tab:combined_table}
\end{table}

\section{Summary and Conclusions}
We have presented a physics informed machine learning (PIML) approach to solving the problem of remaining useful lifetime (RUL) prediction using the C-MAPSS dataset. Our approach, based on stochastic models of the underlying data, is novel in that we do not use the traditional PIML architectures such as PINNs, but rather augment the training data using estimated underlying quantities. Finally, our apprach was shown to allow for generative synthesis of data (i.e. physics aware synthetic data generation). The statistical distribution of the sensor data was seen to exhibit multi-modal behavior, and our approach is valid in this situation as well.  Our experiments clearly indicate the increased accuracy (measured both using MSE and absolute error) over traditional deep learning (LSTM) approaches. It is expected that future work in this area would involve increasing the scale and scope of physics informed ML methods for predictive diagnostics.


\bibliography{bibli}
\end{document}